\documentclass[11pt,a4paper]{article}

\usepackage[utf8]{inputenc}
\usepackage[T1]{fontenc}

\usepackage[a4paper, margin=25mm]{geometry}

\usepackage{setspace}
\usepackage{amsmath}
\usepackage{amssymb}
\usepackage{amsfonts}

\let\bar\overline

\usepackage{graphicx}
\usepackage{subcaption}
\usepackage{placeins}

\usepackage{booktabs}
\usepackage{multirow}
\usepackage{makecell}
\usepackage{array}

\usepackage{listings}
\usepackage[numbers,sort&compress]{natbib}

\usepackage{xcolor}
\usepackage[hidelinks]{hyperref}

\title{Explaining RhythmFormer: A Systematic XAI Analysis of Periodic Sparse Attention for Remote Photoplethysmography}

\author{%
  Louis Chen\thanks{%
    Department of Mechanical Engineering, National Cheng Kung University, Tainan, Taiwan.
    ORCID: \href{https://orcid.org/0009-0001-8373-2295}{0009-0001-8373-2295}.
    Email: \texttt{louis.chen@nordlinglab.org}.%
  }%
  \and
  Torbj{\"o}rn E.\,M.\ Nordling\thanks{%
    Department of Mechanical Engineering, National Cheng Kung University, Tainan, Taiwan.
    ORCID: \href{https://orcid.org/0000-0003-4867-6707}{0000-0003-4867-6707}.
    Corresponding author: \texttt{torbjorn.nordling@nordlinglab.org}.%
  }%
}

\date{\today}

\begin{document}
\maketitle

\begin{abstract}
Remote photoplethysmography (rPPG) transformers achieve low heart-rate error on benchmarks, yet their decisions remain opaque ---a growing concern as rPPG moves toward clinical heart rate estimation.
Existing rPPG XAI is dominated by qualitative heatmap inspection without quantitative faithfulness metrics or physiology-grounded validation, leaving a gap between visual plausibility and auditable evidence.
We address this gap with three contributions.
First, we adapt four attribution methods (raw attention, rollout, flow, Beyond Intuition) to RhythmFormer's bi-level routing attention with top-$k$ selection.
Second, we introduce a \emph{skin coverage} metric quantifying how much attribution mass falls on physiologically valid skin regions.
Third, we adapt the SaCo faithfulness coefficient from its original classification setting to rPPG regression by using the MAE between original and perturbed predicted rPPG waveforms as the perturbation impact.
Applying these tools, we quantify a multi-hop leakage effect under sparse top-$k$ routing: attention rollout and flow almost completely restores the connections that individual refined-attention layers explicitly set to zero.
Beyond Intuition mitigates this via its value-projection-weighted rollout and gradient-supported mask, attaining the highest median refined skin coverage ($0.83$ vs.\ $0.57$ for vanilla rollout, i.e., standard attention rollout without value-projection or gradient weighting) and faithfulness ($F{=}0.92$) among the evaluated methods on UBFC-rPPG with RhythmFormer.
We therefore identify Beyond Intuition as the strongest candidate in this setting, while emphasizing that validation across diverse datasets and model variants remains an important direction for future work.
A case study on a low-SaCo outlier further shows all four methods recovering consistently once an artefactual region is replaced, suggesting consistent SaCo behavior across attribution families in this illustrative case.
Together, these two quantitative metrics move XAI for rPPG beyond visual heatmap inspection toward auditable numerical evidence about spatial alignment and perturbation faithfulness---a step toward more trustworthy rPPG XAI.

\end{abstract}


\bigskip

\section{Introduction}
\label{sec:intro}

Traditional vital sign monitoring methods such as Electrocardiography (ECG) and Photoplethysmography (PPG) rely on direct skin contact, which causes discomfort, raises infection risk, and limits applicability in scenarios such as neonatal care and remote monitoring.
Remote Photoplethysmography (rPPG) addresses these limitations by estimating vital signs from subtle skin color changes in standard video recordings~\cite{verkruysse2008remote}, enabling applications such as telemedicine, fitness tracking, and stress monitoring, though robustness against varying illumination, motion artifacts, and skin tone differences remains an open challenge.
The rapid advancement of deep learning has driven rPPG from traditional signal-processing pipelines~\cite{de2013robust,wang2016algorithmic} to increasingly sophisticated neural architectures: 2-dimensional convolutional neural networks (2D CNNs)~\cite{chen2018deepphys,liu2023efficientphys}, 3D CNNs~\cite{yu2019compressed}, and more recently transformers~\cite{yu2022physformer,yu2023physformer,zou2025rhythm,qian2024DualPathTCSS} that exploit long-range spatiotemporal dependencies critical for quasi-periodic rPPG signals.
Among them, RhythmFormer~\cite{zou2025rhythm} reports competitive performance on five benchmarks (PURE~\cite{stricker2014non}, UBFC-rPPG~\cite{bobbia2019unsupervised}, COHFACE~\cite{heusch2017reproducible}, VIPL-HR~\cite{niu2018vipl}, MMPD~\cite{tang2023mmpd}) while keeping a single-stream architecture without auxiliary branches or multi-task heads, making it particularly amenable to XAI analysis.
As rPPG transformers continue to lower heart-rate error on benchmark datasets, their interpretability is gaining attention as a research priority in its own right---before such systems can enter diagnostic workflows, clinicians need to understand and trust the model's reasoning.
Physiological studies have identified specific facial regions, notably the forehead and cheeks, as carrying superior plethysmographic information~\cite{Kwon2015ROI}, but inquiring about the spatial focus of deep learning models to see if it aligns with this knowledge requires dedicated interpretation methods.
XAI techniques are therefore essential not only for clinician trust but also for validating the physiological integrity of rPPG-based decisions.
Yet XAI applications in rPPG remain limited.
Existing approaches rely individually on raw attention~\cite{qian2024DualPathTCSS}, Grad-CAM~\cite{Li2024tscan}, or gradient saliency~\cite{liu2025physkannet,huang2025ddrppg}, without systematic cross-method comparison or quantitative faithfulness evaluation.
Moreover, raw attention's validity as a proxy for model focus is contested~\cite{jain2019attentionexplanation,wiegreffe2019attentionexplanation,serrano2019attentioninterpretable}: a plausible-looking heatmap may leave the prediction unchanged when its highlighted regions are masked, indicating the model relies on features the explanation does not capture.

In this work, we present a comprehensive XAI framework for analyzing RhythmFormer's periodic sparse attention.
Our contributions are:
\begin{itemize}
    \item We adapt four complementary XAI methods---raw attention, attention rollout~\cite{abnar2020attention}, attention flow~\cite{abnar2020attention}, and Beyond Intuition~\cite{chen2023intuition}---to RhythmFormer's bi-level routing attention with sparse top-$k$ selection.
    \item We introduce a skin attention quantification metric measuring the overlap between model attention and physiologically relevant skin regions (face and neck) across all Hierarchical Temporal Periodic Transformer (TPT) levels.
    \item We adapt SaCo~\cite{wu2024saco} to the rPPG regression setting, providing a quantitative faithfulness assessment for rPPG XAI.
    \item We conduct systematic experiments on UBFC-rPPG, examining the clip-level correlation between attribution metrics (skin coverage, SaCo) and prediction quality (waveform Pearson~$r$, heart rate (HR) error) to test whether the model's spatial focus matches the physiologically intuitive expectation that attending to skin yields more accurate estimates.
\end{itemize}

\section{Related Work}
\label{sec:related}

\subsection{XAI Methods}

Explainability methods for transformers can be broadly categorized into three families~\cite{Mersha2024explainable,bhati2024survey}.
\textit{Perturbation-based} methods (e.g., LIME~\cite{ribeiro2016lime}, SHAP~\cite{lundberg2017shap}) probe predictions by input modification, while \textit{gradient-based} methods (e.g., saliency~\cite{simonyan2014deep}, LRP~\cite{bach2015pixel}, Grad-CAM~\cite{Selvaraju2019grad-cam}) compute attribution from the model's gradient signal; among them, Integrated Gradients~\cite{sundararajan2017axiomatic} accumulates gradients along a baseline-to-input path and reappears later as the $F^c$ component of Beyond Intuition (Section~\ref{sec:beyond_intuition}).
\textit{Attention-based} methods directly visualize or aggregate attention weights.
Raw attention provides per-layer views, while attention rollout~\cite{abnar2020attention} recursively multiplies augmented attention matrices to capture transitive token relationships.
Attention flow~\cite{abnar2020attention} models the attention graph as a flow network and computes maximum flow between input and output tokens.
Beyond Intuition~\cite{chen2023intuition} bridges the attention and gradient families by combining attention rollout with integrated gradients, decomposing attribution into attention perception ($P^{(L)}$, how information flows) and reasoning feedback ($F^c$, which attention entries the model is sensitive to).
Since attention is the locus of RhythmFormer's spatial decisions (bi-level routing with top-$k$ selection), we focus on attention-based attribution and adopt four complementary variants---raw, rollout, flow, Beyond Intuition---spanning per-layer, cumulative, and gradient-augmented forms to mitigate the contested faithfulness of raw attention alone~\cite{jain2019attentionexplanation,wiegreffe2019attentionexplanation,serrano2019attentioninterpretable,pruthi2020learning}.

\subsection{XAI in rPPG}

Existing rPPG XAI works---Dual-path TokenLearner~\cite{qian2024DualPathTCSS}, PhysKANNet~\cite{liu2025physkannet}, DD-rPPGNet~\cite{huang2025ddrppg}, and TS-CAN+~\cite{Li2024tscan}---rely on raw attention, gradient saliency, or Grad-CAM for visualization, with TS-CAN+ revealing that the model sometimes responds to non-physiological areas such as red clothing; none provides quantitative faithfulness evaluation.
A related example is CIN-rPPG~\cite{li2024channelwise}, which inspects its own channel-spatial interactive learning maps as a module-internal visualisation rather than applying a post-hoc XAI method.

\subsection{Quantification in XAI---Faithfulness Evaluation}

A common quantitative approach is cumulative perturbation, which progressively removes salient pixels; however, because each step is applied on top of all previous removals, the impact of individual pixel groups is confounded~\cite{wu2024saco}.
SaCo~\cite{wu2024saco} overcomes this by perturbing each group individually and computing pairwise consistency between salience ranking and prediction impact.
To our knowledge, this is the first work to adapt SaCo to rPPG regression.

\section{Method}
\label{sec:method}

\subsection{RhythmFormer Overview}
\label{sec:rhythmformer}

\begin{figure*}[tbh]
\centering
  \includegraphics[width=\textwidth]{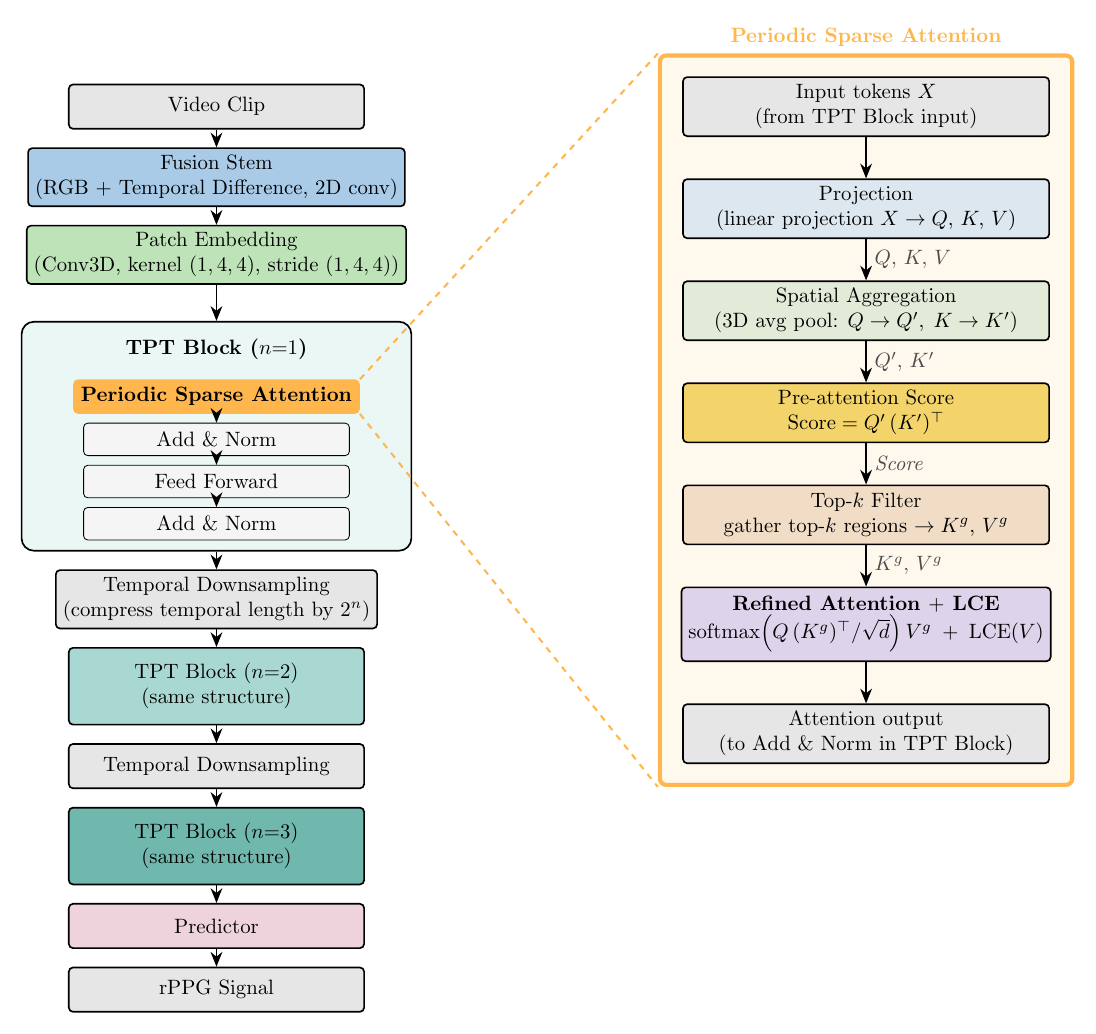}
\caption{RhythmFormer architecture. \textit{Left}:~Overall pipeline consisting of a Fusion Stem, Patch Embedding, three hierarchical TPT levels ($n=1,2,3$) separated by Temporal Downsampling (compressing temporal length by $2^n$ at level $n$), and a Predictor; \textit{Right}:~Periodic Sparse Attention detail within each TPT block: a \textit{pre-attention} stage spatially aggregates $Q, K$ and selects the top-$k$ regions $K^g, V^g$ from coarse scores $Q'(K')^\top$; a \textit{refined attention} stage then applies scaled dot-product attention over the gathered regions with $\mathrm{LCE}(V)$.}
\label{fig:rhythmformer}
\label{fig:rhythmformer_arch}
\label{fig:psa}
\end{figure*}

RhythmFormer~\cite{zou2025rhythm} is an end-to-end transformer for rPPG estimation.
Given an RGB facial video $X \in \mathbb{R}^{3 \times T \times H \times W}$, the model produces an rPPG signal $\hat{y} \in \mathbb{R}^{T}$.
The architecture (Figure~\ref{fig:rhythmformer_arch}) consists of four components:
(1)~a \textbf{Fusion Stem} that integrates raw frames with temporal frame differences to capture BVP color variations;
(2)~\textbf{Patch Embedding} that divides the output into non-overlapping tokens;
(3)~a \textbf{Hierarchical Temporal Periodic Transformer} (TPT) with three hierarchical levels, each containing two TPT blocks (six attention layers in total) employing periodic sparse attention, where level $n \in \{1,2,3\}$ operates at temporal resolution $T/2^n$ (i.e., $80, 40, 20$ frames for a 160-frame clip); and
(4)~a \textbf{Predictor Head} that produces the 1D rPPG waveform.

\paragraph{Periodic sparse attention.}
The key innovation is a two-stage attention mechanism within each TPT block (Figure~\ref{fig:psa}).
In the \textit{pre-attention stage}, queries $Q$ and keys $K$ are projected via temporal difference convolutions (TDC) and subjected to 3D average pooling, yielding $Q'$ and $K'$.
Pre-attention scores $\text{Score} = Q'(K')^\top$ are computed, and top-$k$ selection retains only the highest-scoring regions.
In the \textit{refined attention stage}, the gathered key-value pairs undergo standard scaled dot-product attention with softmax:
\begin{equation}
    SA = \text{Softmax}\!\left(\frac{Q(K^g)^\top}{\sqrt{d}}\right) V^g + \text{LCE}(V),
    \label{eq:refined_attn}
\end{equation}
where $K^g, V^g$ are the gathered key and value tokens selected by the pre-attention stage, and $d$ is the per-head embedding dimension.
The scaling factor $\sqrt{d}$ follows the standard scaled dot-product formulation~\cite{vaswani2017attention} and is used to stabilize training.
LCE (Local Context Enhancement) is a depthwise $3{\times}3{\times}3$ convolution applied to $V$ as a parallel branch; since RhythmFormer omits explicit positional embeddings, LCE provides an implicit positional inductive bias through its local receptive field.
All six Bi-Level Routing Attention (BRA) layers use top-$k$ routing with $k{=}40$.
The candidate region set contains $640$, $640$, and $320$ regions in the three TPT levels, respectively, so top-$40$ retains roughly $6.3\%$, $6.3\%$, and $12.5\%$ of candidate regions per layer.
These values follow the default RhythmFormer configuration~\cite{zou2025rhythm}.

\paragraph{Selection rationale and remaining gap.}
We select RhythmFormer because it reports competitive rPPG performance across multiple benchmarks~\cite{zou2025rhythm} while retaining a single-stream transformer design.
This is important for cumulative attention-based methods (rollout, flow, Beyond Intuition), which trace cross-layer attention chains; unlike architectures with parallel pathways or external modules such as PhysFormer++~\cite{yu2023physformer}, Dual-path TokenLearner~\cite{qian2024DualPathTCSS}, and CodePhys~\cite{chu2025codephys}, RhythmFormer provides a clean sequential chain across its three bi-level routing levels (TPT$_1$, TPT$_2$, TPT$_3$).

\subsection{Attention-Centered Attribution Methods}
\label{sec:attention_methods}

We extract attention matrices from all six attention layers (three TPT levels $\times$ two blocks each) and adapt four attribution methods: raw attention, attention rollout, attention flow, and Beyond Intuition.
We denote the two BRA blocks within TPT level $s$ as TPT$_s$-B0 and TPT$_s$-B1, where B0 is the first BRA block.
Each TPT block produces two distinct attention matrices stemming from RhythmFormer's two-stage attention design: the \textit{coarse pre-attention} scores at region granularity ($4{\times}4$, $N{=}16$) and the \textit{refined attention} weights at token granularity ($8{\times}8$, $N{=}64$, obtained after the top-$k$ selection and softmax).
Since the refined attention is sparse---each query region only attends to the top-$k$ selected key regions---we reconstruct it as a dense $N{\times}N$ matrix with zeros at excluded positions to enable the matrix operations required by rollout, flow, and Beyond Intuition.
Raw attention, rollout, and flow are applied independently to both pre-attention and refined matrices; Beyond Intuition is applied only to the refined matrices (see paragraph below).
All attention matrices are head-averaged before any aggregation, so $A^{(l)}$ in subsequent equations refers to the head-mean attention at layer $l$.

\paragraph{Raw attention.}
For each layer $l$, we compute the received attention of each spatial position $k$ by summing over all query positions: $r_k^{(l)} = \sum_q A^{(l)}_{qk}$.
This column-wise aggregation measures how much each position is attended to as a key, serving as an alternative to the \texttt{[CLS]}-row aggregation used in ViT-based methods~\cite{dosovitskiy2020image}, which is unavailable in RhythmFormer.
Results are grouped by TPT level (averaging within each level) to yield three maps per resolution.

\paragraph{Attention rollout.}
Following~\cite{abnar2020attention}, we recursively multiply augmented attention matrices $\hat{A}^{(l)} = 0.5\,I + 0.5\,A^{(l)}$ across all layers, where the identity term models the residual skip connection in each transformer block that bypasses attention with roughly equal weight:
$\tilde{A} = \hat{A}^{(L)} \cdots \hat{A}^{(1)}$.
Since the three TPT levels operate at different temporal resolutions ($T{=}80, 40, 20$ for clip length 160), we downsample all attention matrices to the coarsest resolution ($T{=}20$) by temporal mean pooling: each token at the higher-resolution level is mapped to its corresponding bin at $T{=}20$ (group sizes $4$ and $2$ for levels $1$ and $2$), and entries falling into the same $(q, k)$ bin pair are averaged.
Since RhythmFormer has no \texttt{[CLS]} token, we aggregate via column-sum over query positions, which corresponds to input-side saliency under our top-down multiplication convention (rows index query/output side, columns index key/input side): $\psi_k = \sum_q \tilde{A}_{qk}$.

\paragraph{Attention flow.}
Following~\cite{abnar2020attention}, we model attention as a flow network where each attention weight $A^{(l)}_{ts}$ acts as the capacity of an edge from token $s$ at layer $l{-}1$ to token $t$ at layer $l$, and compute maximum flow between all input--output token pairs.
For each input token $i$ at layer $0$, we sum the max-flow values to every output token $j$ at layer $L$:
$\phi_i = \sum_j \text{MaxFlow}(v_i^{(0)}, v_j^{(L)})$.
This yields the total information capacity that flows out of each input position $i$, providing a source-side saliency complementary to the column-sum aggregation used for raw attention and rollout.

\paragraph{Beyond Intuition.}
\label{sec:beyond_intuition}
Beyond Intuition~\cite{chen2023intuition} decomposes attribution as
\begin{equation}
    T = P^{(L)} \odot F^c,
    \label{eq:bi}
\end{equation}
where $\odot$ denotes the Hadamard product.
$P^{(L)}$ is the W-corrected attention rollout.
Here, vanilla rollout denotes the standard attention-rollout baseline before BI's corrections: it multiplies attention matrices across layers and implicitly equates the attention weight with each token's downstream influence, which is equivalent to assuming the value projection $W$ is the identity.
In practice, every token is transformed into $z_j W$ before being passed forward, so a token with large attention weight but small $\|z_j W\|$ propagates little information.
We therefore scale each token's column in the rollout by $\alpha_j = \|z_j W\| / \|z_j\|$ (where $W$ denotes the effective value projection $W_{\text{out}} W_v$), down-weighting tokens whose projected magnitude is small relative to their input magnitude and removing the information-flow distortion that vanilla rollout ignores.
$F^c$ is the integrated gradient of the loss w.r.t.\ the last-layer attention:
\begin{equation}
    F^c = \text{ReLU}\!\left(\frac{1}{N_{\text{steps}}}\sum_{s=1}^{N_{\text{steps}}} \nabla_{A^{(L)}} \mathcal{L}\!\left(\frac{s}{N_{\text{steps}}} \cdot X\right)\right),
    \label{eq:fc}
\end{equation}
where gradients are averaged over $N_{\text{steps}}$ interpolation steps from a zero baseline to the input, mitigating gradient saturation; ReLU retains only positions whose increase raises the loss.
The Hadamard product combines structural importance (from rollout) with decision-relevant gradient signal.
Following the original formulation~\cite{chen2023intuition}, which computes $F^c$ from the last transformer layer's attention, we apply BI exclusively to the refined attention matrices ($8{\times}8$) where the model's spatial decisions are made, rather than the pre-attention routing scores ($4{\times}4$).
We adapt BI to RhythmFormer with four modifications:
\begin{itemize}
    \item \emph{Sparse-to-dense reconstruction.} The W-correction preserves the top-$k$-excluded zeros since $\alpha_j \cdot 0 = 0$, so positions the model does not attend to remain zero throughout $P^{(L)}$.
    \item \emph{Temporal resolution unification.} As in rollout, all layers are downsampled to $T{=}20$ to enable cross-level matrix multiplication in $P^{(L)}$.
    \item \emph{Mean-intensity baseline.} Since inputs are z-score normalized per channel and per clip, the effective baseline at $k{=}0$ is a uniform video at the per-channel per-clip mean (matching the perturbation baseline used by SaCo, Section~\ref{sec:saco}) rather than the conventional all-zero (black) baseline. This is semantically more appropriate for rPPG: an all-black baseline would push the network out of distribution, whereas the per-channel mean baseline removes BVP-related variation while staying within the training distribution.
    \item \emph{LCE branch excluded from $W$.} We fold only the linear value-projection chain $W = W_{\text{out}} W_v$ into the W-correction and leave the LCE branch outside $W$. Two reasons: (i) LCE is a depthwise $3{\times}3{\times}3$ convolution rather than a token-wise linear map, so it cannot be expressed as a single matrix multiplication and does not fit the linear-projection form on which the W-correction is derived~\cite{chen2023intuition}; (ii) the W-correction is intended to measure how much each \emph{token's content} contributes to the output via the value pathway, while LCE applies a level-fixed spatial kernel that is not content-routed and therefore makes an approximately uniform contribution to relative token importance. Folding LCE into $W$ would require a convolution-aware extension of BI's derivation, which we leave to future work.
\end{itemize}

\subsection{Faithfulness Evaluation via SaCo}
\label{sec:saco}

We adopt SaCo~\cite{wu2024saco} to evaluate explanation faithfulness.
Given a saliency map, the input is partitioned into $K$ equally-sized groups ranked by salience.
Let $s(G)$ denote the mean saliency score over pixels in group $G$.
For each pair $(G_i, G_j)$ with $s(G_i) \geq s(G_j)$, SaCo checks whether masking $G_i$ causes greater prediction impact than masking $G_j$.
The faithfulness coefficient is:
\begin{equation}
    F = \frac{\sum_{i<j} w_{ij}}{\sum_{i<j} |w_{ij}|} \in [-1, 1],
    \label{eq:saco}
\end{equation}
where $w_{ij} = +(s(G_i) - s(G_j))$ when masking $G_i$ causes greater impact than masking $G_j$ (consistent with the saliency ordering), and $w_{ij} = -(s(G_i) - s(G_j))$ when the impact ordering is inverted.
The signed weight is proportional to the saliency gap, so pairs with larger gaps contribute more to $F$.
Thus $F{=}1$ indicates perfect faithfulness (all pairs consistent) and $F{=}-1$ indicates full inversion; the absolute value $|F|$ measures the degree of correlation between saliency ranking and prediction impact~\cite{wu2024saco}.

\paragraph{Adaptation for regression.}
We replace the classification confidence drop with prediction distance:
\begin{equation}
    \nabla\!\operatorname{pred}(x, G_i) = \text{MAE}\!\left(\Phi(x),\; \Phi(x_{G_i})\right),
    \label{eq:saco_mae}
\end{equation}
where $\Phi(x)$ is the model's predicted rPPG waveform and $x_{G_i}$ denotes $x$ with the pixels in $G_i$ replaced by the per-clip per-channel mean intensity (following SaCo's Algorithm~1, which keeps the perturbed input within the training distribution while removing the spatial content of $G_i$).
A larger MAE indicates stronger model reliance on $G_i$.
Using $\Phi(x)$ as the reference, rather than the ground-truth label, preserves the key property of SaCo~\cite{wu2024saco}: impact is determined solely by model sensitivity, without conflating faithfulness with prediction accuracy.

\subsection{Skin Attention Quantification}
\label{sec:skin_quant}

Since the rPPG signal originates from blood-volume-induced color variations on exposed skin, attention placed off-skin cannot carry pulsatile information; we therefore hypothesize that higher skin coverage in the attention map should be associated with lower prediction error.
To test this, we quantify the overlap between attention heatmaps and skin regions using a pre-trained BiSeNet face parser~\cite{yu2018bisenet}.
To bring the attention and skin mask to a common spatial grid, the raw attention heatmap (at $4{\times}4$ or $8{\times}8$ depending on the method, see Section~\ref{sec:attention_methods}) is min-max normalized to $[0, 1]$ and bicubic-upsampled to the face image resolution $H{\times}W{=}128{\times}128$, yielding $\mathbf{h} \in [0, 1]^{H \times W}$.
The binary skin mask $\mathbf{m} \in \{0, 1\}^{H \times W}$ encodes each pixel as skin (1) or non-skin (0).
The skin coverage metric is:
\begin{equation}
    \text{Skin Coverage} = \frac{\sum_{i} \mathbf{h}_i \cdot \mathbf{m}_i}{\sum_{i} \mathbf{h}_i},
    \label{eq:skin_coverage}
\end{equation}
where $i$ indexes all $H{\times}W$ pixels.
The numerator is the attention mass falling on skin and the denominator is the total attention mass, so the ratio reports the fraction of attention mass concentrated on skin regions.
A value of 1.0 indicates all attention falls on skin; 0.0 indicates no overlap.

\subsection{Waveform-Level Signal Quality}
\label{sec:waveform_quality}

To assess whether attention patterns relate to prediction quality at the signal level, we measure the per-clip Pearson correlation coefficient between the predicted and ground-truth rPPG waveforms:
\begin{equation}
    r = \frac{\sum_{t=1}^{T}(\hat{y}_t - \bar{\hat{y}})(y_t - \bar{y})}{\sqrt{\sum_{t=1}^{T}(\hat{y}_t - \bar{\hat{y}})^2 \cdot \sum_{t=1}^{T}(y_t - \bar{y})^2}},
    \label{eq:pearson_r}
\end{equation}
where $\hat{y}$ and $y$ are the predicted and ground-truth rPPG waveforms within a single clip of $T$ frames.
This waveform-level metric captures signal reconstruction fidelity and is distinct from the HR-level Pearson $\rho$ in Table~\ref{tab:reproduction}, which measures agreement of scalar heart rate estimates across subjects.

\subsection{Heart-Rate-Level Error}
\label{sec:hr_mae}

For the clip-level HR analysis in the XAI correlation figures, we estimate scalar heart rate from the predicted and ground-truth rPPG waveforms following the FFT-based evaluation protocol for RhythmFormer~\cite{zou2025rhythm}.
Each XAI point corresponds to one UBFC-rPPG clip with $T{=}160$ frames at $f_s{=}30$ Hz, a $5.33$ s window.
Both waveforms are detrended ($\lambda{=}100$), bandpass filtered to $0.75$--$2.5$ Hz ($45$--$150$ bpm), and converted to a power spectral density using Welch's method.
The heart rate is the dominant spectral peak in this band:
\begin{equation}
    \operatorname{HR}(y)
    = 60\,\operatorname*{arg\,max}_{f \in [0.75,\,2.5]} \operatorname{PSD}_y(f).
    \label{eq:hr_fft}
\end{equation}
The clip-level HR error is
\begin{equation}
    e_{\operatorname{HR}} =
    \left|\operatorname{HR}(\hat{y}) - \operatorname{HR}(y)\right|,
    \label{eq:hr_error}
\end{equation}
The quantity used for each clip-level correlation point is this absolute HR error.
When aggregating over a set of clips, HR MAE is the mean of $e_{\operatorname{HR}}$ across clips.

\section{Experiments}
\label{sec:experiments}

\subsection{Experimental Setup}
\label{sec:setup}

\paragraph{Dataset.}
We evaluate on UBFC-rPPG~\cite{bobbia2019unsupervised}, which contains 42 videos at $640 \times 480$ resolution and 30~fps, each approximately one minute long (${\sim}1{,}800$ frames), recording subjects in a stationary state.
Following the original RhythmFormer protocol~\cite{zou2025rhythm}, the first 30 samples are used for training and the remaining 12 for testing.
Each test video is segmented into non-overlapping 160-frame clips (\texttt{chunk\_num}~$=\lfloor\text{frame\_count}/160\rfloor$); the trailing segment shorter than 160 frames is discarded, yielding $n{=}141$ clips across the 12 test subjects (12 clips per subject for 11 of the 12 subjects; one subject with a shorter recording yields 9).
We focus on UBFC-rPPG for two reasons.
First, this work is a method paper aimed at establishing an XAI framework rather than benchmarking across datasets, and UBFC's controlled recording conditions reduce dataset-side confounders (motion artefacts, illumination shifts, parsing failures) so that method-level effects such as multi-hop leakage are not entangled with dataset noise.
Second, UBFC-rPPG is the dataset on which the original RhythmFormer paper~\cite{zou2025rhythm} reports results, allowing a direct reproduction check against the published numbers (Sec.~\ref{sec:reproduction}) before any XAI analysis is performed.

\paragraph{Training.}
The model is trained for 30 epochs with batch size 4, learning rate $9 \times 10^{-3}$, and clip length 160 frames on 4 NVIDIA RTX A5000 GPUs.
The hybrid loss is defined as:
\begin{equation}
    \mathcal{L} = 0.2 \cdot \mathcal{L}_{\text{time}} + \mathcal{L}_{\text{freq}},
\end{equation}
where $\mathcal{L}_{\text{time}} = 1 - \rho(\hat{y}, y)$ is the negative Pearson correlation and $\mathcal{L}_{\text{freq}} = \text{CE}(\text{PSD}(\hat{y}), \text{PSD}(y))$ is the cross-entropy over power spectral densities~\cite{zou2025rhythm}.
Data augmentation includes random temporal resampling conditioned on heart rate and random horizontal flips.
No modifications are made to the original RhythmFormer architecture or hyperparameters.

\paragraph{XAI configuration.}
Attention matrices are extracted from all six BRA layers (three TPT levels $\times$ two blocks).
Beyond Intuition uses $N_{\text{steps}}{=}20$ integration steps, following the lower bound recommended by Sundararajan et al.~\cite{sundararajan2017axiomatic} for approximating the integral within 5\%.
SaCo evaluation uses $K{=}8$ spatial groupings with the MAE impact metric; this is a practical trade-off given that the $8{\times}8$ refined-attention grid bounds the maximum number of spatial groupings at 64, and larger $K$ increases computational cost with diminishing evaluation granularity.

\paragraph{Rollout reconstruction analysis.}
To characterize how top-$k$ masking interacts with cumulative rollout, we evaluated the refined $64{\times}64$ attention matrices on all $n{=}141$ test clips.
For each matrix, the analysis uses the $4032$ off-diagonal entries ($64^2{-}64$) and excludes the diagonal because it is preserved by the $\tfrac{1}{2}I$ residual term in the augmented rollout matrices.
For each off-diagonal token pair, we record whether it is zeroed by top-$k$ selection in at least one refined-attention layer and compare its final rollout value against an always-retained baseline consisting of entries that remain non-zero in all six refined-attention layers.

\subsection{Training Reproduction}
\label{sec:reproduction}

Table~\ref{tab:reproduction} compares our reproduced training results against the metrics reported in the original paper~\cite{zou2025rhythm}.
The close agreement confirms that our reproduction is faithful and the model checkpoint is reliable for subsequent XAI analysis.
The slight discrepancies are expected: non-deterministic CUDA reductions under \texttt{DistributedDataParallel} cause the optimization trajectory to diverge across hardware configurations~\cite{pytorch2024Reproducibility}, and because the original protocol uses no independent validation set---selecting the best checkpoint directly on the 12 test subjects via the HR bias term $B_{HR}$~\cite{zou2025rhythm}---this run-to-run variability propagates directly into the reported metrics.

\begin{table}[!htb]
\centering
\caption{Comparison of our reproduced results with the original paper on UBFC-rPPG. $\downarrow$ = lower is better, $\uparrow$ = higher is better.}
\label{tab:reproduction}
\small
\begin{tabular}{lcccc}
\toprule
 & MAE$\downarrow$ & RMSE$\downarrow$ & MAPE$\downarrow$ & $\rho$$\uparrow$ \\
\midrule
Paper~\cite{zou2025rhythm} & 0.50 & 0.78 & 0.47 & 0.99 \\
Ours & 0.45 & 0.76 & 0.46 & 0.997 \\
\bottomrule
\end{tabular}
\end{table}

\section{Results}
\label{sec:results}

\subsection{Qualitative Visualization}
\label{sec:qualitative}

We first present qualitative visualizations to provide intuition for how different XAI methods interpret the model's attention.

\begin{figure*}[tbh]
\centering
\includegraphics[width=\textwidth]{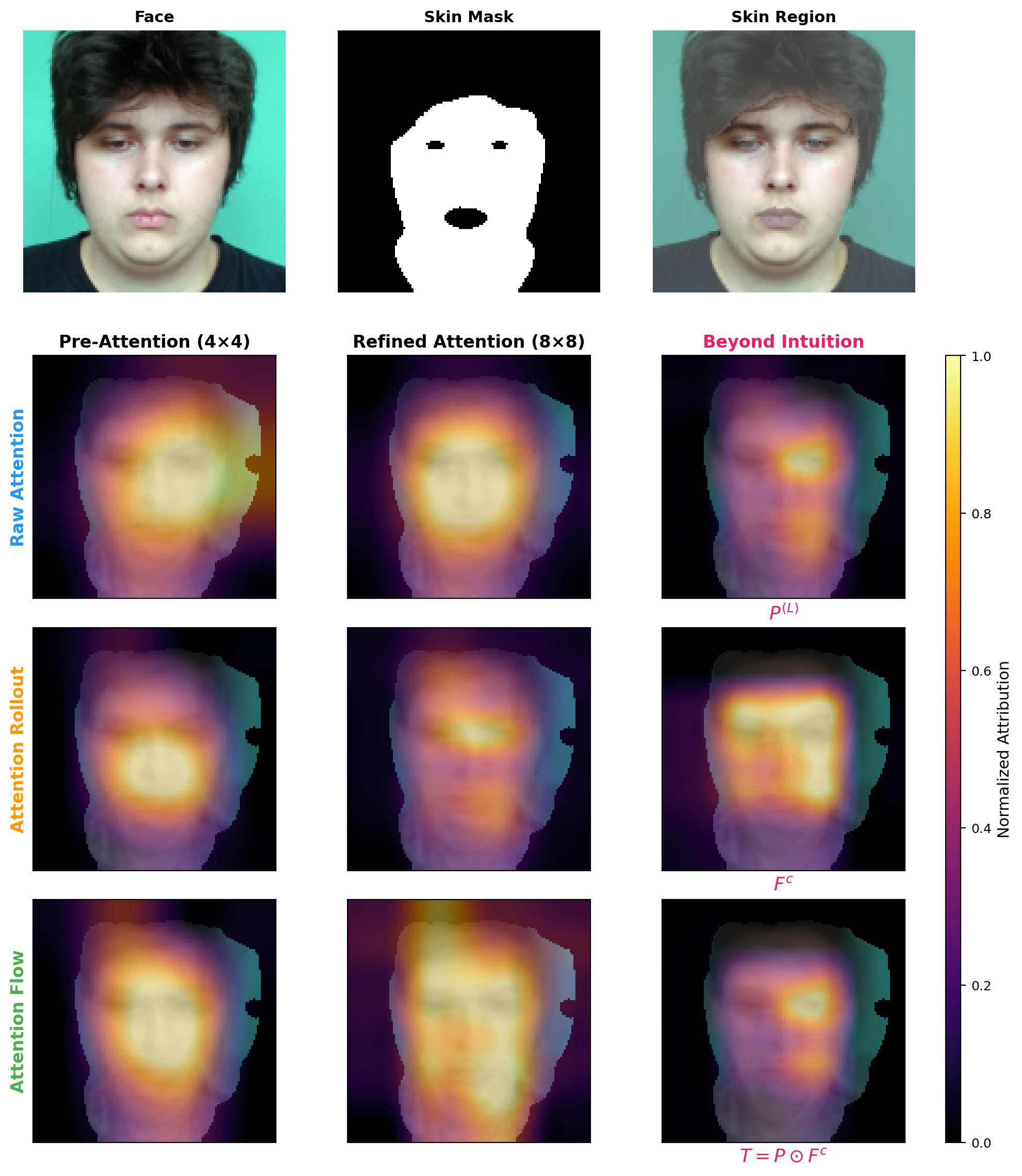}
\caption{Pre-attention vs.\ refined attention with Beyond Intuition on UBFC-rPPG. Top row: representative sample face, skin mask, and skin region overlay. Middle $3 \times 2$ grid: rows = raw attention, rollout, and attention flow; columns = pre-attention ($4{\times}4$) routing scores and refined ($8{\times}8$) post-top-$k$ weights, cross-subject and cross-TPT averaged. Right column: Beyond Intuition components $P^{(L)}$, $F^c$, and $T = P^{(L)} \odot F^c$, cross-subject averaged. Heatmaps are per-cell min-max normalized.}
\label{fig:cross_heatmap}
\end{figure*}

Figure~\ref{fig:cross_heatmap} shows attention heatmaps averaged across all 12 test subjects and three TPT levels, revealing several patterns.
First, all methods consistently highlight the facial skin region rather than background or non-skin areas.
Second, comparing the two columns of attention-based methods reveals the effect of top-$k$ selection: for raw attention, the refined column is visibly more concentrated on skin than the pre-attention column, directly reflecting the sparsification introduced by bi-level routing.
However, this concentration difference is less pronounced for the cumulative methods.
For rollout, the cumulative matrix product can restore attribution to positions excluded by individual layers' top-$k$; attention flow has the same path-based caveat because it aggregates capacity over the same multi-layer attention graph.
We quantify the resulting coverage pattern below and discuss the rollout mechanism and its implication for flow in Section~\ref{sec:discussion}.
Third, Beyond Intuition provides a complementary perspective: its final attribution $T = P^{(L)} \odot F^c$ is notably more concentrated than the attention-based methods, reflecting the sparsity inherited from top-$k$ selection (Section~\ref{sec:discussion}).

Rollout and Beyond Intuition concentrate saliency on the cheeks and forehead---regions with high subcutaneous blood flow~\cite{Kwon2015ROI}---whereas raw attention and attention flow spread more broadly.
This qualitative ordering is quantified in the skin coverage analysis below.

\subsection{Skin Coverage Analysis}
\label{sec:skin_results}

To move beyond individual examples, we quantify the overlap between attention heatmaps and physiologically relevant skin regions.
Using the binary skin mask defined in Section~\ref{sec:skin_quant}, we compute the skin coverage ratio defined in Eq.~\ref{eq:skin_coverage}.
Since each BRA layer in RhythmFormer produces two independent attention matrices---pre-attention weights ($Q'K'^\top$, region-level) and refined attention weights (post-top-$k$ softmax, token-level)---we compute skin coverage separately at each resolution: pre-attention at $4{\times}4$ and refined attention at $8{\times}8$.
For cumulative methods (rollout, attention flow), each version operates on the corresponding set of matrices independently.

\begin{figure}[tbh]
\centering
\includegraphics[width=\textwidth]{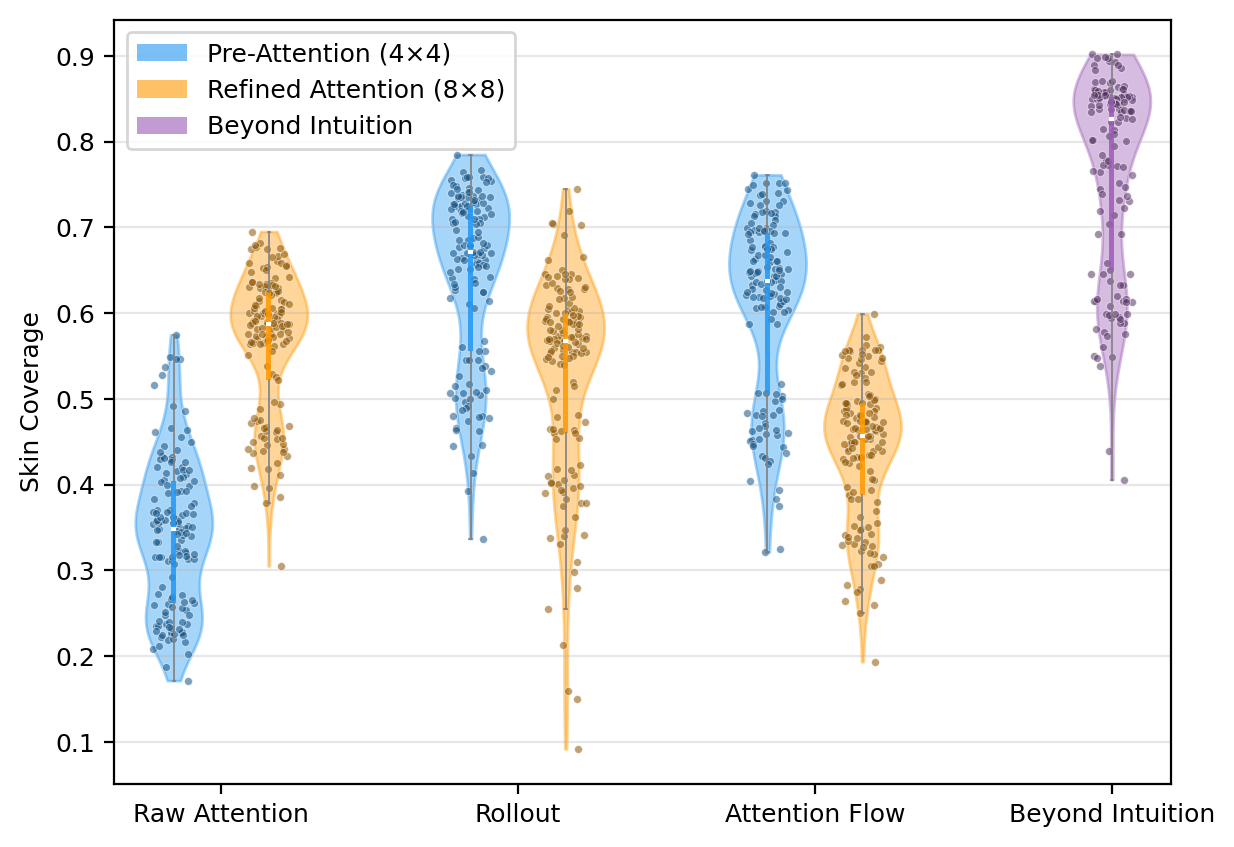}
\caption{Violin plots of skin coverage for all four XAI methods. Pre-attention (4$\times$4) and refined attention (8$\times$8) are shown separately for attention-based methods. Beyond Intuition appears only at 8$\times$8 because it operates on the refined attention matrices (Section~\ref{sec:beyond_intuition}): the pre-attention routing scores lack softmax normalization, making their gradient signal nearly uniform and uninformative for attribution.}
\label{fig:skin_violin}
\end{figure}

Figure~\ref{fig:skin_violin} compares the skin coverage distributions across all four methods and all test subjects.
For each attention-based method (raw attention, rollout, attention flow), each data point represents the mean skin coverage of a single clip, averaged across the three TPT levels.
Three patterns emerge.
First, for raw attention, the refined stage achieves significantly higher skin coverage than the pre-attention stage ($\Delta = +0.22$, $p < 0.001$).
Second, for rollout and attention flow, this relationship reverses: the refined stage yields significantly \textit{lower} skin coverage than the pre-attention stage ($\Delta = -0.12$ and $-0.17$, both $p < 0.001$).
Third, Beyond Intuition achieves significantly higher skin coverage than all attention-based methods at the refined level ($p < 0.001$), but with markedly greater inter-subject variability---its distribution spans from below 0.4 to above 0.8.
We analyze the mechanisms underlying these three phenomena in Section~\ref{sec:discussion}.
Table~\ref{tab:summary} reports the median skin coverage alongside median SaCo faithfulness for each method.

\begin{table}[!htb]
\centering
\caption{Median skin coverage and SaCo faithfulness ($F$, $K{=}8$) on UBFC-rPPG. For attention-based skin coverage, each clip is first averaged across the three TPT levels, and the table reports the median across clips. $\uparrow$ = higher is better. Best in \textbf{bold}. Beyond Intuition operates only at the refined level (Section~\ref{sec:beyond_intuition}).}
\label{tab:summary}
\small
\begin{tabular}{lccc}
\toprule
 & Skin Cov.\ (pre)$\uparrow$ & Skin Cov.\ (refined)$\uparrow$ & SaCo $F$$\uparrow$ \\
\midrule
Raw Attention    & 0.35 & 0.59 & 0.85 \\
Rollout          & \textbf{0.67} & 0.57 & 0.85 \\
Attention Flow   & 0.64 & 0.46 & 0.84 \\
Beyond Intuition & ---  & \textbf{0.83} & \textbf{0.92} \\
\bottomrule
\end{tabular}
\end{table}

\subsection{Rollout Reconstruction of Top-\texorpdfstring{$k$}{k}-Masked Connections}
\label{sec:rollout_reconstruction}

Top-$k$ selection explicitly zeros many refined-attention entries in individual layers.
For example, TPT$_1$-B0 zeros $2036/4032=50.5\%$ of off-diagonal entries, and the six refined-attention layers contain $4197$ such zero-entries on average over the $141$ clips.
The six-layer rollout restores every one of these excluded entries to a non-zero value: $R^{(L)}$ is fully dense, so $100\%$ of the top-$k$-masked off-diagonal entries are recovered through multi-hop paths.
This effect is not limited to the full six-layer traversal.
Within a single TPT level, $98$--$100\%$ of the first block's top-$k$ zeros become non-zero after the two-layer rollout of that level.
The recovered values are also not negligible: the median rollout value at recovered entries is $0.0106$, only $1.8{\times}$ smaller than at always-retained entries ($0.0192$), and the $99$th percentile of recovered values ($0.0392$) exceeds the always-retained median.
Overall, recovered entries account for approximately $48\%$ of plain rollout's total off-diagonal mass.

\subsection{Faithfulness Evaluation}
\label{sec:faithfulness}

\begin{figure}[tbh]
\centering
\includegraphics[width=\textwidth]{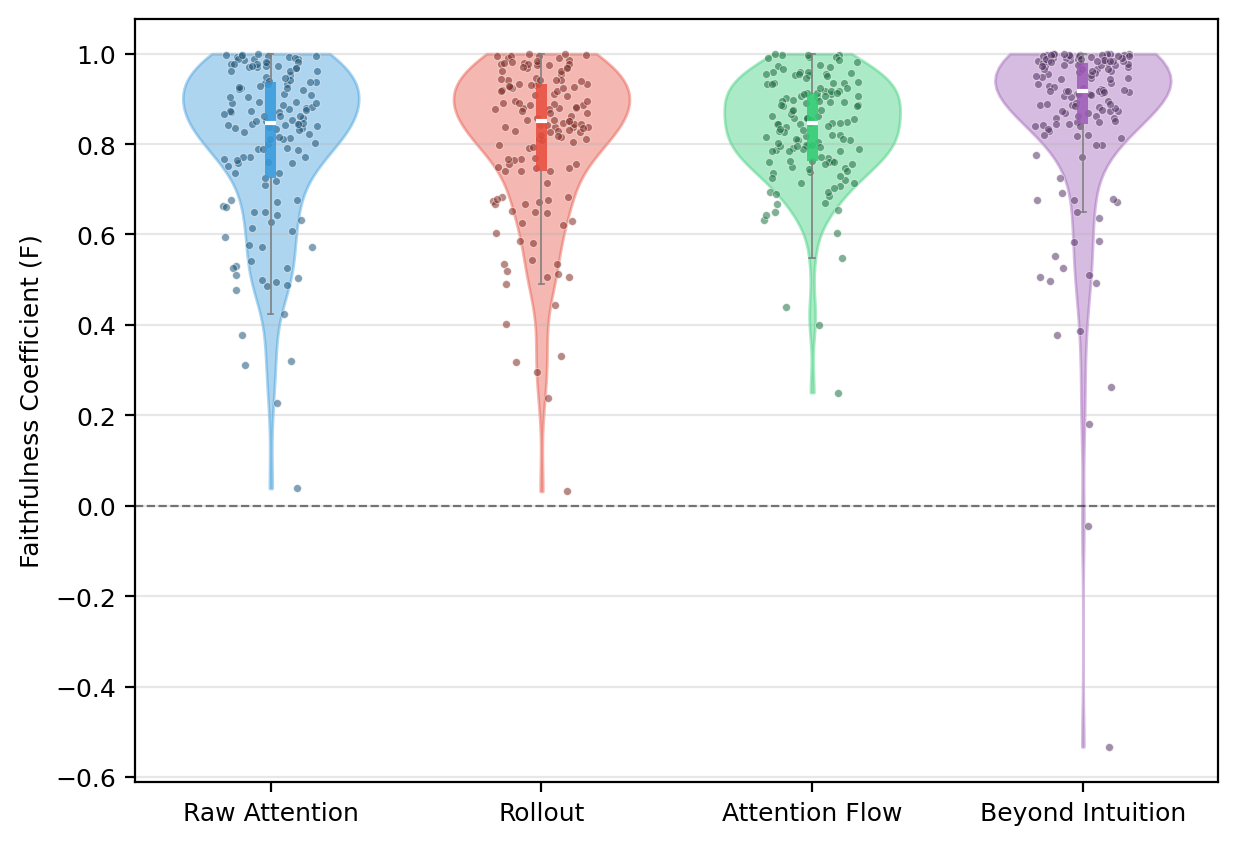}
\caption{Violin plots of SaCo faithfulness scores by method across all test subjects. Beyond Intuition achieves the highest median faithfulness; the distributional shapes are broadly comparable across methods.}
\label{fig:saco_violin}
\end{figure}

Figure~\ref{fig:saco_violin} shows the distribution of SaCo faithfulness scores across all test clips (median values in Table~\ref{tab:summary}).
Beyond Intuition achieves the highest overall faithfulness, but notably exhibits several clips with negative SaCo scores; other methods also contain low-scoring outliers at the subject level.
We examine these outlier cases in Section~\ref{sec:discussion}.

\subsection{Clip-Level Correlation Analysis}
\label{sec:clip_correlation}

The preceding sections evaluate XAI metrics in aggregate; here we examine clip-level relationships between two prediction-quality metrics---HR-level mean absolute error (MAE; FFT-based) and waveform-level Pearson~$r$ (time-domain)---and two XAI metrics---skin coverage and SaCo faithfulness---using Spearman's rank correlation coefficient.
Each point represents a single 160-frame clip; points are colored by subject to visualize within- versus between-subject variability.

\begin{figure}[tbh]
\centering
\includegraphics[width=\textwidth]{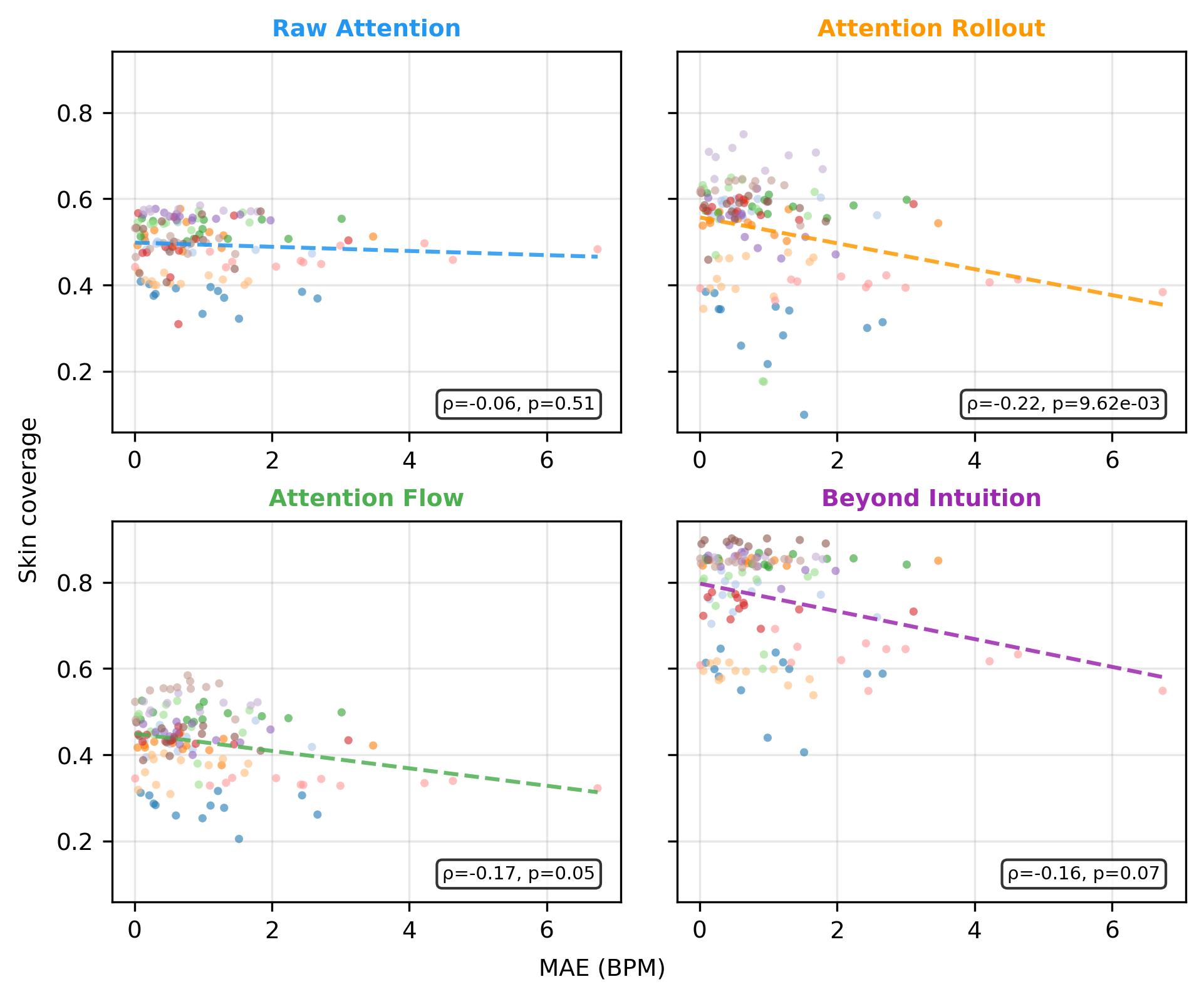}
\caption{MAE vs.\ skin coverage per clip for each XAI method ($n{=}141$ clips across $12$ test subjects). Each point represents one 160-frame clip; colors denote different test subjects. Spearman correlations: rollout $\rho{=}{-}0.22$ ($p{=}0.009$), flow $\rho{=}{-}0.17$ ($p{=}0.05$), Beyond Intuition $\rho{=}{-}0.16$ ($p{=}0.07$), raw $\rho{=}{-}0.06$ ($p{=}0.51$).}
\label{fig:mae_skin}
\end{figure}

\paragraph{MAE vs.\ skin coverage.}
If skin coverage reflects the model attending to physiologically informative regions, we expect a negative correlation between skin coverage and MAE.
Figure~\ref{fig:mae_skin} matches this expected direction across all four methods, although the correlations are weak: only attention rollout reaches significance, while attention flow and Beyond Intuition are borderline.
This indicates that HR accuracy is shaped by factors beyond spatial attention to skin.

\begin{figure}[tbh]
\centering
\includegraphics[width=\textwidth]{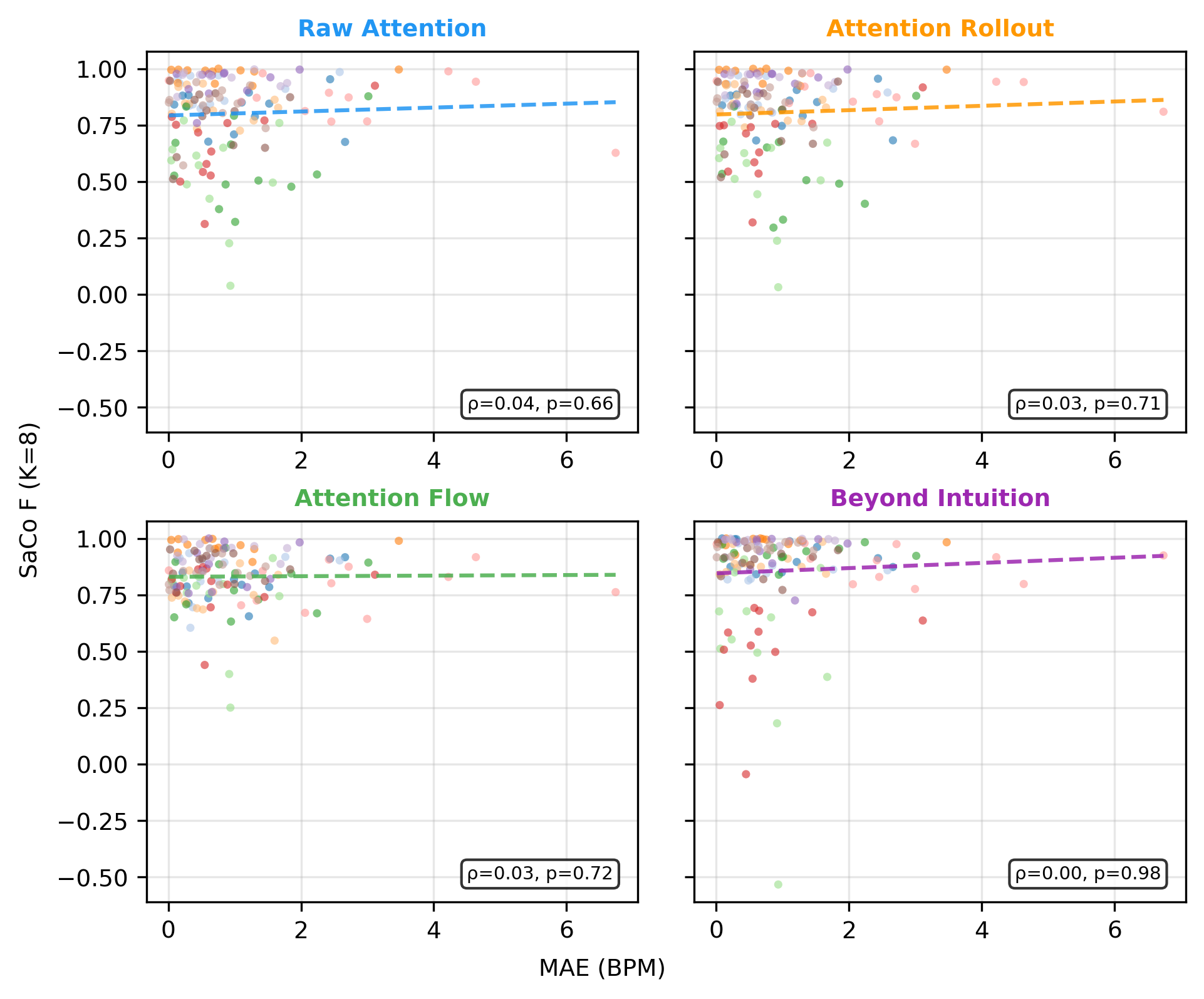}
\caption{MAE vs.\ SaCo faithfulness ($F$, $K{=}8$) per clip for each XAI method ($n{=}141$ clips across $12$ test subjects). Each point represents one 160-frame clip; colors denote different test subjects. Spearman correlations: all $|\rho| \leq 0.04$, all $p \geq 0.66$ (Beyond Intuition $\rho{=}0.00$, $p{=}0.98$).}
\label{fig:mae_saco}
\end{figure}

\paragraph{MAE vs.\ SaCo faithfulness.}
We expect lower MAE to coincide with higher SaCo, i.e., clips on which the model predicts well should also be those on which its attribution is more faithful to its decision.
Figure~\ref{fig:mae_saco} does not exhibit this expected trend: no method shows a significant correlation.
We discuss the underlying reasons in Section~\ref{sec:discussion}.

\begin{figure}[tbh]
\centering
\includegraphics[width=\textwidth]{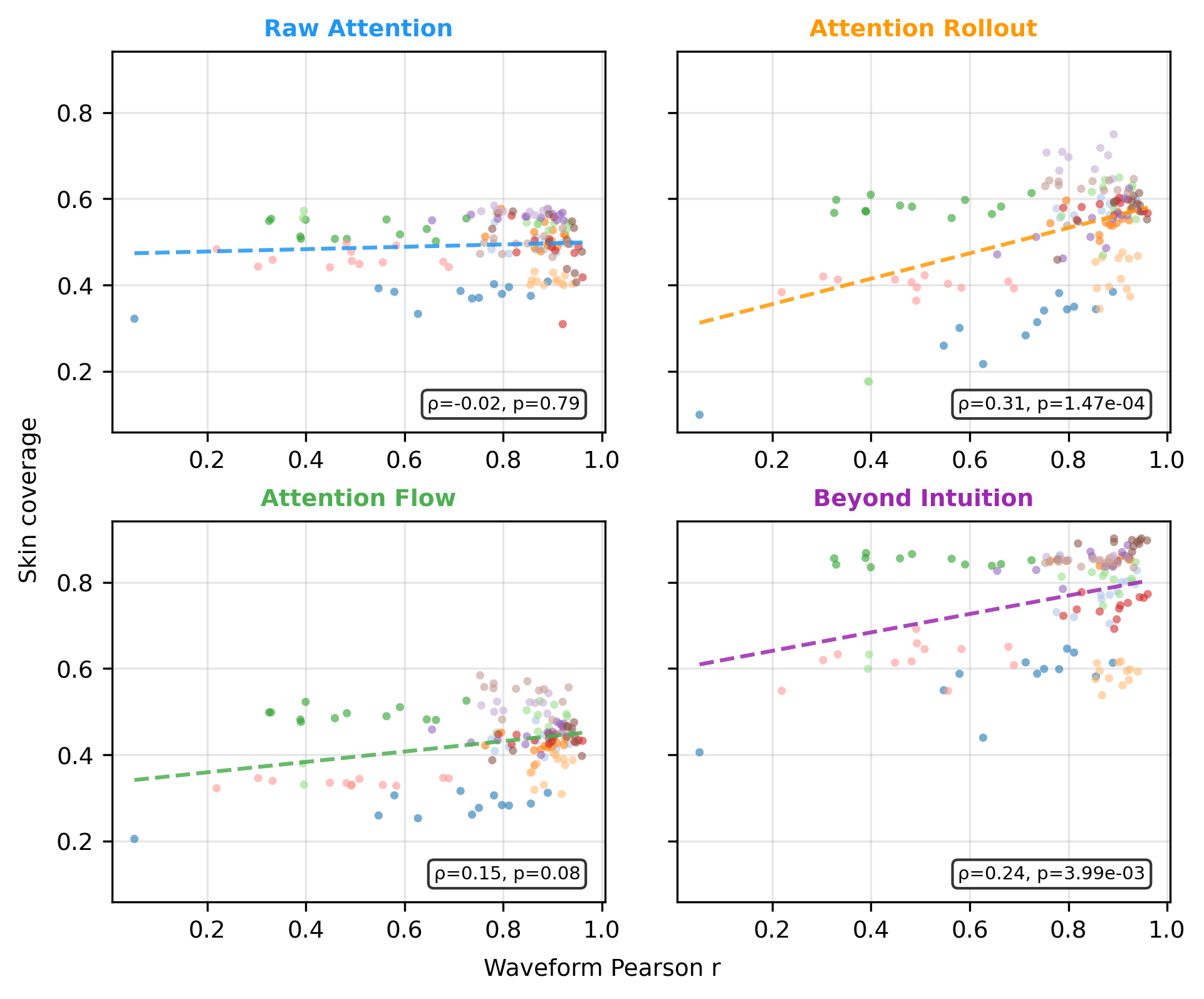}
\caption{Waveform quality (Pearson~$r$) vs.\ skin coverage per clip for each XAI method ($n{=}141$ clips across $12$ test subjects). Higher Pearson~$r$ indicates better waveform reconstruction. Colors denote different test subjects. Spearman correlations: rollout $\rho{=}0.31$ ($p<0.001$), Beyond Intuition $\rho{=}0.24$ ($p{=}0.004$), flow $\rho{=}0.15$ ($p{=}0.08$), raw $\rho{=}{-}0.02$ ($p{=}0.79$).}
\label{fig:pearson_skin}
\end{figure}

\paragraph{Waveform quality vs.\ skin coverage.}
To disentangle waveform reconstruction quality from heart rate accuracy, we replace the MAE axis with the per-clip Pearson~$r$ (Eq.~\ref{eq:pearson_r}; Figure~\ref{fig:pearson_skin}); we expect higher skin coverage to coincide with higher Pearson~$r$, i.e., better waveform fidelity.
Figure~\ref{fig:pearson_skin} partially matches this expectation: attention rollout and Beyond Intuition show significant positive correlations, while attention flow and raw attention do not reach significance.
Notably, these positive correlations are stronger than the corresponding MAE--coverage correlations in Figure~\ref{fig:mae_skin}, indicating that skin attention tracks waveform fidelity more closely than HR accuracy.

\paragraph{Asymmetric impact of skin attention: waveform vs.\ HR.}
Both clip-level findings are consistent with the expectation that skin attention contributes to prediction quality: skin coverage shows a significant positive correlation with waveform fidelity (Figure~\ref{fig:pearson_skin}; rollout $\rho{=}0.31$, Beyond Intuition $\rho{=}0.24$, both $p < 0.005$), and a directionally negative trend with HR error across all four methods (Figure~\ref{fig:mae_skin}).
These two observations correspond to successive stages of the same propagation chain: spatial attention to skin yields cleaner BVP waveforms, which in turn yield sharper spectral peaks and lower HR error.
The first stage (skin $\to$ waveform) is directly visible in our data; the second stage (waveform $\to$ HR) is attenuated under UBFC-rPPG's controlled recording conditions, where sufficiently high SNR allows even imperfect waveforms to produce unambiguous spectral peaks, saturating HR accuracy (MAE~$\approx 0$ for most subjects) and masking the influence of spatial attention at the HR level.
Although this leaves only a weak negative MAE--skin correlation in the present data, we would expect this correlation to strengthen substantially in noisier datasets (motion artifacts, variable illumination, lower SNR), where the spectral margin between the HR peak and noise floor narrows and waveform-quality differences propagate more directly to HR error.
Under this framing, skin coverage remains a meaningful XAI metric: it captures where the model directs spatial attention and correlates with prediction quality in the expected directions, with HR-level effects masked here only by UBFC's saturated conditions.

\subsection{Artifact Ablation Case Study}
\label{sec:artifact_ablation_results}

\begin{figure}[tbh]
    \centering
    \includegraphics[width=\textwidth]{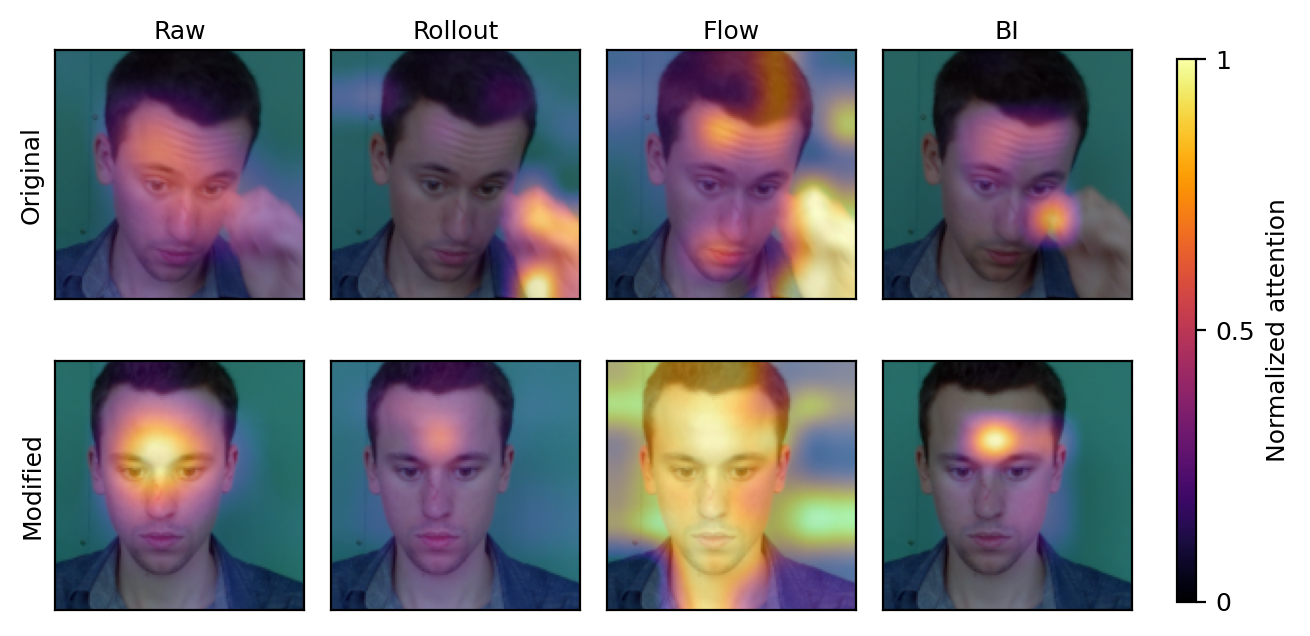}
    \caption{Hand occlusion ablation on subject~46 clip~11. Top: original; bottom: with the hand-occlusion frames replaced by stable neighbouring frames from the same clip. Across all four attribution methods (Raw, Rollout, Flow, BI), the replacement recovers a more skin-aligned distribution; BI SaCo lifts from $-0.53$ to $0.76$ on this clip.}
    \label{fig:artifact_ablation_main}
    \end{figure}

To complement the aggregate analyses, we report one low-SaCo outlier as a result-level case study; the interpretation is deferred to Section~\ref{sec:discussion}.
Subject~46 clip~11 has the lowest Beyond Intuition SaCo in the test set ($F=-0.53$), and frame-by-frame inspection of the raw video showed that the subject rubs their eye for $6$ out of $160$ frames ($3.75\%$ of the input).
We replaced these six hand-occlusion frames with stable neighbouring frames from the same clip---preserving the $160$-frame input length and the temporal context---and re-ran the full forward pass and XAI pipeline, so that only those $6$ frames change.
Figure~\ref{fig:artifact_ablation_main} compares the original and ablated heatmaps, and Table~\ref{tab:artifact_ablation} reports SaCo $F$ and skin coverage before and after the intervention.
Across all four attribution methods, both SaCo and skin coverage recover after the intervention; the per-method values are reported in Table~\ref{tab:artifact_ablation}.
\begin{table}[!t]
\centering
\small
\caption{Artefact ablation on subject~46 clip~11. SaCo faithfulness $F$ and skin coverage  before and after replacing the $6$ hand-artefact frames ($3.75\%$ of the clip) with stable neighbouring frames from the same clip. Skin coverage uses the same BiSeNet skin mask (computed from the original frame~$0$) for both conditions; raw, flow, and BI are reported as their per-clip average, and rollout uses the refined average. All four methods' SaCo recover toward $F\!\approx\!0.8$ (Beyond Intuition shows the largest gain), while skin coverage moves in the same direction with smaller magnitudes.}
\label{tab:artifact_ablation}
\begin{tabular}{lcccccc}
\toprule
& \multicolumn{3}{c}{SaCo $F$} & \multicolumn{3}{c}{Skin coverage } \\
\cmidrule(lr){2-4}\cmidrule(lr){5-7}
Method & Orig & Mod & $\Delta$ & Orig & Mod & $\Delta$ \\
\midrule
Raw attention      & $0.038$  & $0.835$ & $+0.797$ & $0.542$ & $0.621$ & $+0.080$ \\
Rollout            & $0.032$  & $0.802$ & $+0.770$ & $0.113$ & $0.390$ & $+0.277$ \\
Attention flow     & $0.146$  & $0.864$ & $+0.718$ & $0.238$ & $0.328$ & $+0.090$ \\
Beyond Intuition   & $-0.533$ & $0.760$ & $+1.293$ & $0.698$ & $0.761$ & $+0.063$ \\
\bottomrule
\end{tabular}
\end{table}
\FloatBarrier

\section{Discussion}
\label{sec:discussion}

\subsection*{Mechanism: how bi-level routing shapes attribution}

Section~\ref{sec:rollout_reconstruction} shows that plain rollout reconstructs top-$k$-masked refined-attention connections into non-zero multi-hop weights, with recovered entries contributing approximately $48\%$ of rollout's off-diagonal mass.
This multi-hop leakage explains the skin-coverage reversal observed for rollout in Section~\ref{sec:skin_results}: even though top-$k$ at every layer removes a large fraction of weight from non-skin positions, the rollout product carries roughly half of the off-diagonal mass back to excluded positions, so refined rollout coverage drops \emph{below} pre-attention ($\Delta{=}{-}0.12$, Fig.~\ref{fig:skin_violin}).
Attention flow shows the same empirical refined-vs-pre drop ($\Delta{=}{-}0.17$), and it is mechanistically exposed to the same sparse-routing caveat because max-flow is computed over paths in the same layered attention graph; however, we do not quantify its leakage mass here and therefore do not claim that its magnitude matches rollout.
Visually the rollout heatmap still looks face-centred---the \emph{peak} values remain on skin, dominated by always-retained attention---but the spread of recovered mass to non-skin pixels is large enough to drag the skin-coverage ratio down.
Importantly, this leakage hurts \emph{spatial} interpretability but not \emph{faithfulness}: rollout and attention flow remain comparable to raw attention on SaCo ($0.85$ and $0.84$ vs.\ $0.85$, Tab.~\ref{tab:summary}), indicating that their cumulative/path-based attributions still fall on regions the model genuinely relies on, even when those regions extend beyond the skin mask.

Beyond Intuition (BI)~\cite{chen2023intuition} forms its attribution as $T=P^{(L)}\!\odot\!F^c$, where $P^{(L)}$ is the W-corrected rollout (reweighting each token by its value-projection magnitude $\alpha_j=\|z_j W\|/\|z_j\|$) and $F^c$ is a last-layer integrated-gradient mask.
We compute skin coverage on every test clip ($n{=}141$, face\_neck mask) and report the median across all clips for each component of the BI pipeline (Tab.~\ref{tab:bi_decomposition}).
Replacing vanilla rollout with the W-corrected rollout $P^{(L)}$ raises median skin coverage from $0.57$ to $0.68$ ($\Delta{=}{+}0.11$; \textbf{$141/141$ clips improved}).
Because every clip improves with no exception, the data directly show that the W-correction step attenuates the multi-hop leakage characterized above.
The integrated-gradient mask $F^c$ does not admit the same kind of direct comparison: it is constructed from a last-layer gradient rather than from a chain of sparse attention matrices, so there is no ``uncorrected $F^c$'' to serve as a fair baseline.
($F^c$ on its own has median skin coverage $0.66$, but this value cannot be interpreted as the effect of an isolated correction step.)
Its effect is instead observed in combination with $P^{(L)}$: the Hadamard product $T=P^{(L)}\!\odot\!F^c$ raises median skin coverage further from $0.68$ to $0.83$ ($\Delta{=}{+}0.15$; \textbf{$141/141$ clips improved}).

Taken together, the two steps lift median skin coverage by $+0.28$ over vanilla rollout, with every clip improving at every step.
Beyond Intuition is therefore not just nominally more faithful---it explicitly closes the multi-hop leakage gap that plain rollout suffers from, indicating that suppressing this leakage is the better attribution strategy under top-$k$ sparse attention.

\begin{table}[t]
\centering
\footnotesize
\caption{BI decomposition: per-clip skin coverage ($n{=}141$ clips across $12$ test subjects, face\_neck mask). Each pairwise comparison has $141/141$ clips improving in the same direction.}
\label{tab:bi_decomposition}
\begin{tabular}{lr}
\toprule
Comparison & $\Delta$ median \\
\midrule
$P^{(L)}$ vs vanilla rollout & $+0.15$ \\
$T$ vs $P^{(L)}$              & $+0.15$ \\
$T$ vs $F^c$                  & $+0.17$ \\
\bottomrule
\end{tabular}
\end{table}

\subsection*{Spatial attention does not uniformly track prediction quality}

SaCo faithfulness shows no significant correlation with either HR error (Figure~\ref{fig:mae_saco}) or waveform quality, indicating that attribution consistency under perturbation captures an aspect of model behavior independent of prediction accuracy at both the waveform and heart rate levels.
This independence does not diminish SaCo's utility: it evaluates whether the attribution map correctly identifies which spatial regions the model relies on, as verified by perturbation, regardless of whether the model's prediction is accurate.
A clip can be poorly predicted yet faithfully attributed---the model's attention is genuinely concentrated on those regions, they simply do not carry a strong rPPG signal---or accurately predicted with low faithfulness, where the correct prediction arises from distributed attention that no single region dominates.

\paragraph{XAI variability under saturated performance.}
Even among accurately predicted clips, both skin coverage and SaCo faithfulness vary substantially rather than converging to uniformly high values.
This indicates that strong predictive performance does not imply a uniform internal strategy: two clips with equally accurate heart rate predictions may rely on different spatial attention patterns---a distinction invisible to aggregate performance metrics.

\subsection*{Outlier analysis: artifact ablation}

The artefact ablation reported in Section~\ref{sec:artifact_ablation_results} (Fig.~\ref{fig:artifact_ablation_main} and Table~\ref{tab:artifact_ablation}) is a single-clip empirical observation; here we discuss what it implies for SaCo as a cross-method metric.
The intervention was triggered by Beyond Intuition's strongly negative score on this clip ($F=-0.53$): a negative $F$ means the regions BI flags as most salient receive \emph{lower} prediction impact than the regions it deems unimportant---attention anti-correlated with what actually drives the output---which made the clip a candidate for manual inspection rather than dismissal as random noise.
Once the six hand-artefact frames are replaced, all four attribution methods recover toward $F\!\approx\!0.8$, and because the same direction of change emerges across raw attention, rollout, attention flow, and Beyond Intuition---four algorithmically distinct attributions---a method-specific artefact explanation is unlikely; the four methods agree both that this clip is problematic and on the direction of change after the intervention, suggesting that SaCo behaves consistently across attribution families on this clip.
Skin coverage moves in the same direction but is less responsive on this clip, consistent with the broader observation that skin coverage and SaCo capture related but non-identical aspects of attribution quality.
We treat this as illustrative rather than confirmatory: a single-clip ablation cannot establish that SaCo measures genuine attention/impact alignment in general, and a systematic multi-outlier study would be required to support that stronger claim.

\paragraph{Limitations.}
Two caveats apply.
\textbf{(i)} Our evaluation uses UBFC-rPPG, which has only $12$ test subjects under controlled conditions (stable indoor lighting, near-frontal pose, no motion artefacts on most clips).
RhythmFormer's HR error saturates near zero on these clips, so the dataset is not complex enough to test whether spatial attention metrics (skin coverage, SaCo) actually correlate with prediction quality in the regimes where the model is challenged.
The weak MAE--skin correlation we observe could equally reflect an absence of relationship or a saturation effect that masks one---the present data cannot distinguish these.
\textbf{(ii)} Our analysis covers only UBFC-rPPG and only RhythmFormer.
Whether the multi-hop leakage pattern reported here is specific to RhythmFormer's top-$k$ schedule, or generalises to other sparse-attention rPPG models (e.g., PhysFormer variants, BiFormer-based pipelines on other tasks), remains an open question; patterns observed here should not be promoted to general claims about sparse vision transformers without cross-model replication.

\paragraph{Future directions.}
Several extensions follow naturally.
First, replication on more challenging benchmarks (e.g., PURE~\cite{stricker2014non}, MMPD~\cite{tang2023mmpd}), where motion artefacts and variable illumination push the model out of UBFC saturation, would let us test whether skin coverage and SaCo track prediction quality once HR error is non-trivial.
Second, our integrated-gradient baseline for $F^c$ is currently the per-clip mean-intensity reference; alternative baselines (e.g., black, Gaussian noise, or learned) may shift $F^c$ and consequently $T$, and a systematic comparison would clarify how baseline choice affects the attribution.
Third, RhythmFormer's three TPT levels process features at different spatial and temporal scales; computing BI attribution per TPT level rather than only at the last layer could reveal whether each level emphasises distinct physiological features---for example, whether earlier levels capture coarse skin-region attention and later levels refine to specific pulsatile regions.

\section{Conclusion}
\label{sec:conclusion}

This work proposes a quantitative comparison framework for four representative XAI attribution methods on transformer-based rPPG models---raw attention, attention rollout, attention flow, and Beyond Intuition---evaluated through two complementary metrics: skin coverage over canonical face regions and SaCo faithfulness.
Applied to RhythmFormer on UBFC-rPPG, the framework yields findings consistent with prior physiological expectations on the spatial side: skin coverage shows a moderate positive correlation with waveform-level Pearson and a weak negative trend with HR MAE.
Both directions are consistent with the hypothesis that spatial attention to skin contributes to prediction quality, although the correlation strengths remain modest---the weaker MAE branch in particular may reflect UBFC's saturated HR regime, though the present data alone cannot rule out a genuine absence of relationship.
On the faithfulness side, although SaCo is not strongly correlated with MAE on a per-clip basis, an artefact-ablation case study on a low-SaCo outlier (subject~46 clip~11) shows that SaCo recovers in the same direction across all four attribution methods once the offending frames are replaced; we read this as suggestive that SaCo behaves consistently across attribution families on this clip, while noting that a single case cannot rule out method-specific biases in general.

Connecting these findings to the literature, our results challenge the dense-attention assumption underlying cumulative attention attribution under sparse routing.
For attention rollout~\cite{abnar2020attention}, we directly show that under RhythmFormer's top-$k$ bi-level scheme~\cite{zhu2023biformer}, multi-hop paths re-fill the entries that individual layers explicitly zero out---what we term multi-hop leakage---so positions filtered at one or more layers still accumulate roughly half of rollout's off-diagonal attribution mass.
Attention flow~\cite{abnar2020attention} should be interpreted with the same caution because it also aggregates over the multi-layer attention graph via max-flow, although the present analysis does not quantify its leakage magnitude separately.
Beyond Intuition sidesteps the rollout-specific issue: the value-projection-weighted rollout attenuates the leaked mass, and the integrated-gradient mask restricts attribution to gradient-supported regions.
We further extend Beyond Intuition from its original classification setting~\cite{chen2023intuition} to regression by treating the MAE between original and perturbed predicted rPPG waveforms as the SaCo~\cite{wu2024saco} impact signal, paralleling the way the original method used confidence drops under classification.

These observations suggest two directions.
First, on UBFC-rPPG Beyond Intuition achieves the highest median skin coverage and SaCo among the four methods while remaining robust to multi-hop leakage, although with greater inter-clip variability than rollout and attention flow---including a small number of clips with markedly lower scores. On this dataset and model we therefore see Beyond Intuition as a strong candidate for transformer-based rPPG attribution under sparse routing, while noting that its higher tail risk may make rollout or flow preferable when stable per-clip behaviour matters more than typical performance, and that this preference still needs to be tested on more datasets and models before being promoted to a general recommendation.
Second, the present evaluation is bounded by UBFC-rPPG's saturated HR regime; replicating this framework on benchmarks with motion artefacts and variable illumination~\cite{stricker2014non,tang2023mmpd} should sharpen the MAE--skin-coverage relationship that is dampened here.

\section*{Acknowledgments}
This work was supported by the Taiwan National Science and Technology Council (NSTC) under Grant 111-2221-E-006-186 and Grant 114-2221-E-006-089.

\section*{Declaration on Generative AI}
During the preparation of this work, the authors used Claude (Anthropic) in order to: Grammar and spelling check, code generation for figure plotting scripts, and LaTeX formatting assistance.
After using this tool, the authors reviewed and edited the content as needed and take full responsibility for the publication's content.

\bibliographystyle{unsrtnat}
\bibliography{references}

\end{document}